%% file: main-3135-Jain.tex
\documentclass[11pt,a4paper]{article}
\usepackage{times,latexsym}
\usepackage{url}
\usepackage[T1]{fontenc}

\usepackage[acceptedWithA]{tacl2018v2}

\usepackage{xspace,mfirstuc,tabulary}

\newif\iftaclinstructions
\taclinstructionsfalse 
\iftaclinstructions
\renewcommand{\confidential}{}
\renewcommand{\anonsubtext}{(No author info supplied here, for consistency with
TACL-submission anonymization requirements)}
\newcommand{\instr}
\fi

\iftaclpubformat 

\else

\fi

\usepackage{times}
\usepackage{latexsym}
\usepackage{hyperref}       
\usepackage{url}            
\usepackage{amsfonts}       
\usepackage{amsmath,amssymb,amsthm}
\usepackage{nccmath}

\usepackage{subcaption,siunitx,booktabs}
\usepackage{placeins}

\usepackage{nicefrac}       
\usepackage{microtype}      
\usepackage{lipsum}
\usepackage{caption}
\usepackage{enumitem}
\usepackage{pifont}
\usepackage{wrapfig}
\usepackage{textcomp}
\usepackage{float}
\usepackage{graphicx}
\usepackage{longtable}
\usepackage[normalem]{ulem}
\usepackage{makecell}
\usepackage{arydshln}
\usepackage{xspace}
\usepackage{soul}
\usepackage{color}
\usepackage{multirow, booktabs}
\usepackage{bbding}
\usepackage{booktabs}
\usepackage{tabularx}
\usepackage{tcolorbox}
\usepackage{pgf,tikz}
\usepackage{tikz-qtree}
\usepackage{array}
\usepackage{tikz} 
\usepackage{stackengine}
\usepackage[]{algorithm2e}
\usetikzlibrary{arrows,decorations.markings}
\usetikzlibrary{arrows.meta}

\DeclareFontShape{OT1}{cmtt}{bx}{n}
{
<5> <6> <7> <8> <9>
<10> <10.95> <12> <14.4> <17.28> <20.74> <24.88> cmbtt10
}{}
\DeclareFontShape{OT1}{cmtt}{b}{n}
{<->sub * cmtt/bx/n}{}

\usepackage{pifont}
\definecolor{lightblue}{rgb}{.90,.95,1}

\definecolor{c1}{RGB}{178,24,43}
\definecolor{c2}{RGB}{239,138,98}
\definecolor{c3}{RGB}{253,219,199}
\definecolor{c4}{RGB}{209,229,240}
\definecolor{c5}{RGB}{103,169,207}
\definecolor{c6}{RGB}{33,102,172}

\title{Memory-Based Semantic Parsing}

\author{Parag Jain  \textnormal{and} Mirella Lapata\\
Institute for Language, Cognition and Computation\\
School of Informatics, University of Edinburgh\\
 10 Crichton Street, Edinburgh EH8 9AB\\
\texttt{parag.jain@ed.ac.uk}~~~~\texttt{mlap@inf.ed.ac.uk}\\
}

\date{}
\newcommand\blfootnote[1]{%
  \begingroup
  \renewcommand\thefootnote{}\footnote{#1}%
  \addtocounter{footnote}{-1}%
  \endgroup
}

\begin{document}
\maketitle
\blfootnote{This is a pre-MIT Press publication version.}
\input{1_abstract}
\input{2_introduction}
\input{6_related_work}
\input{3_model}
\input{4_results}
\input{5_analysis}
\input{7_conclusion}
\FloatBarrier

\bibliography{anthology.bib, tacl2018.bib}
\bibliographystyle{acl_natbib}

\end{document}

%% file: 1_abstract.tex
\begin{abstract}
  We present a memory-based model for context-dependent semantic
  parsing. Previous approaches focus on enabling the decoder to copy
  or modify the parse from the previous utterance, assuming there is a
  dependency between the current and previous parses. In this work, we
  propose to represent contextual information using an external
  memory. We learn a context memory controller that manages the memory
  by maintaining the cumulative meaning of sequential user
  utterances. We evaluate our approach on three semantic parsing
  benchmarks.  Experimental results show that our model can better
  process context-dependent information and demonstrates improved
  performance without using task-specific decoders.
\end{abstract}


%% file: 2_introduction.tex
\section{Introduction}

Semantic parsing is the task of converting natural language utterances
into machine interpretable meaning representations such as executable
queries or logical forms. It has emerged as an important component in
many natural language interfaces \citep{10.1145/3318464.3383128} with
applications in robotics~\citep{dukes-2014-semeval}, question
answering~\citep{zhong2018seqsql,yu-etal-2018-spider}, dialogue
systems \cite{artzi-zettlemoyer-2011-bootstrapping}, and the Internet
of Things~\citep{10.1145/3038912.3052562}.

Neural network based approaches have led to significant improvements
in semantic parsing \citep{zhong2018seqsql, kamath2019a,
  yu-etal-2018-spider, yavuz-etal-2018-takes, yu-etal-2018-syntaxsqlnet} across domains and
semantic formalisms.  The majority of existing studies focus on
parsing utterances in isolation, and as a result they cannot readily
transfer in more realistic settings where users ask multiple
inter-related questions to satisfy an information need.
In this work, we study  \emph{context-dependent} semantic parsing
focusing specifically on  text-to-SQL
generation, which has emerged as a popular application  area in recent
years.

\begin{figure*}[t]
\begin{itemize}[label={}]
    \item Q1: \textcolor{red}{What Continental flights} \textcolor{green}{go from Chicago} \textcolor{blue}{to Seattle} \textcolor{cyan}{before 10 am} \textcolor{magenta}{in morning 1993 February twenty sixth}
    \item SQL1: \textcolor{red}{( SELECT DISTINCT flight.flight\_id FROM flight WHERE ( flight.airline\_code = 'CO'} AND \textcolor{green}{( flight . from\_airport IN ( SELECT airport\_service . airport\_code FROM airport\_service WHERE airport\_service . city\_code IN ( SELECT city . city\_code FROM city WHERE city.city\_name = 'CHICAGO' ))} AND \textcolor{blue}{( flight . to\_airport IN ( SELECT airport\_service . airport\_code FROM airport\_service WHERE airport\_service . city\_code IN ( SELECT city . city\_code FROM city WHERE city.city\_name = 'SEATTLE' ))} AND \textcolor{cyan}{( flight.departure\_time $<$ 1000} ) ) ) )   ) ; 
\item Q2: \textcolor{red}{Continental flights} \textcolor{cyan}{before noon} \textcolor{gray}{that have a meal}
\item Q3: \textcolor{red}{Continental flights} \textcolor{cyan}{before 2 pm}
\item Q4: \textcolor{magenta}{On 1993 February twenty seventh}
\item Q5: All Continental flights leaving Chicago before 8 am on 1993 February twenty seventh
\end{itemize}
\caption{Example utterances from a user interaction in the ATIS
  dataset. Utterance segments referring to the same entity or objects are
  in same color. SQL queries corresponding to Q2--Q5 follow a
  pattern similar to Q1 and are not  shown for the sake of brevity.}
\label{colored_interaction}
\end{figure*}

Figure~\ref{colored_interaction} shows a sequence of utterances in an
interaction. The discourse focuses on a specific \emph{topic} serving
a specific information need, namely finding out which Continental
flights leave from Chicago on a given date and time. Importantly,
interpreting each of these utterances, and mapping them to a database
query to retrieve an answer needs to be situated in a particular
context as the exchange proceeds.  The topic further evolves as the
discourse transitions from one utterance to the next and constraints
(e.g., TIME or PLACE) are added or revised.  For example, in Q2 the
TIME constraint \textit{before 10am} from Q1 is revised to
\textsl{before noon}, and in Q3 to \textsl{before 2pm}.  Aside from
such \emph{topic extensions} \citep{chai-jin-2004-discourse}, the
interpretation of Q2 and Q3 depends on Q1, as it is implied that the
questions concern Continental flights that go from Chicago to
Seattle, not just any Continental flights, however the phrase
\textsl{from Chicago to Seattle} is elided from Q2 and Q3. The
interpretation of Q4 depends on Q3 which in turn depends on
Q1. Interestingly, Q5 introduces information with no dependencies on
previous discourse and in this case, relying on information from
previous utterances will lead to incorrect SQL queries.

The problem of contextual language processing has been most widely
studied within dialogue systems where the primary goal is to
incrementally fill pre-defined slot-templates, which can be then used
to generate appropriate natural language responses
\citep{INR-074}. But the rich semantics of SQL queries makes the task
of contextual text-to-SQL parsing substantially different. Previous
approaches \citep{suhr-etal-2018-learning, zhang-etal-2019-editing}
tackle this problem by enabling the decoder to copy or modify the
\emph{previous} queries under the assumption that they contain all
necessary context for generating the current SQL query. The utterance
history is encoded in a hierarchical manner and although this is a
good enough approximation for most queries (in existing datasets), it
is not sufficient to model long-range discourse
phenomena~\citep{grosz-sidner-1986-attention}.

Our own work draws inspiration from Kintsch and van Dijk's
\shortcite{kintsch_toward_1978} text comprehension model. In their
system the process of comprehension involves three levels of
operations. Firstly, smaller units of meaning, i.e., propositions, are
extracted and organized into a coherent whole (\emph{microstructure});
some of these are stored in a working memory buffer and allow to
decide whether new input overlaps with already processed
propositions. Secondly, the gist of the whole is condensed
(\emph{macrostructure}). And thirdly, the previous two operations
generate new texts in working with the memory. In other words, the
(short and long term) memory of the reader gives meaning to the text
read. They propose three macro rules, viz., deletion, generalization,
and construction as essential to reduce and organize the detailed
information of the microstructure of the text.  Furthermore, previous
knowledge and experience are central to the interpretation of text
enabling the reader to fill information gaps.

Our work borrows several key insights from \citet{kintsch_toward_1978}
without being a direct implementation of their model.  Specifically,
we also break down input utterances into smaller units, namely
phrases, and argue that this information can be effectively utilized
in maintaining contextual information in an interaction. Furthermore,
the notion of a \emph{memory} buffer which can be used to store and
process new and old information plays a prominent role in our
approach. We propose a \textbf{Mem}ory-based \textbf{C}ont\textbf{E}xt
model (which we call MemCE for short) for keeping track of contextual
information, and learn a context memory controller that manages the
memory. Each interaction (sequence of user utterances) maintains its
context using a memory matrix. User utterances are segmented into a
sequence of phrases representing either new information to be added
into the memory (e.g., \textit{that have a meal} in
Figure~\ref{colored_interaction}) or old information which might
conflict with current information in memory and needs to be updated
(e.g.,~\textit{before 10 am} should be replaced with \textit{before
  noon} in Figure~\ref{colored_interaction}).  Our model can
inherently add new content to memory, read existing content by
accessing the memory, and update old information.


We evaluate our approach on the
ATIS~\citep{suhr-etal-2018-learning,dahl-etal-1994-expanding},
SParC~\citep{yu-etal-2019-sparc}, and CoSQL~\citep{yu-etal-2019-cosql}
datasets. We observe performance improvements when we combine MemCE
with existing models underlying the importance of more specialized
mechanisms for processing context information. In addition, our model
brings interpretability in how the context is processed.  We are able
to inspect the learned memory controller and analyze whether important
discourse phenomena such as coreference and ellipsis are modeled.

%% file: 6_related_work.tex
\section{Related Work}

Sequence-to-sequence neural networks \cite{bahdanau2014neural} have
emerged as a general modeling framework for semantic parsing,
achieving impressive results across different domains and semantic
formalisms
(\citealt{dong-lapata-2016-language,jia-liang-2016-data,iyer-etal-2017-learning,wang-etal-2020-rat,zhong2018seqsql,yu-etal-2018-spider},
\emph{inter alia}). The majority of existing work has focused on
mapping natural language utterances into machine-readable meaning
representations \emph{in isolation} without utilizing context
information. While this is useful for environments consisting of
one-shot interactions of users with a system (e.g.,~running QA queries
on a database), many settings require extended interactions between a
user and an automated assistant (e.g.,~booking a flight). This
makes the one-shot parsing model inadequate for many scenarios.

In this paper we are concerned with the lesser studied problem of
\emph{contextualized} semantic parsing where previous utterances are
taken into account in the interpretation of the current
utterance. Earlier work
\citep{miller-etal-1996-fully,zettlemoyer-collins-2009-learning,ijcai2017-571}
has focused on symbolic features for representing context, e.g.,~by
explicitly modeling discourse referents, or the flow of
discourse. More recent neural methods extend the sequence-to-sequence
architecture to incorporate contextual information either by modifying
the encoder or the decoder. Context-aware encoders resort to
concatenating the current utterance with the utterances preceding it
\citep{suhr-etal-2018-learning,zhang-etal-2019-editing} or focus on
the history of the utterances most relevant to the current decoder
state \citep{ijcai2020-495}.  The decoders take context
representations as additional input and often copy segments from the
previous query
\citep{suhr-etal-2018-learning,zhang-etal-2019-editing}. Hybrid
approaches
\cite{iyyer-etal-2017-search,guo-etal-2019-towards,ijcai2020-495,
  Lin2019GrammarbasedNT} employ neural networks for representation
learning but use a grammar for decoding (e.g., a sequence of actions
or an intermediate representation).



A tremendous amount of work has taken place in the context of
discourse modeling focusing on extended texts
\citep{mann1988rhetorical, hobbs1985coherence} and
dialogue~\citep{grosz-sidner-1986-attention}. \citet{kintsch_toward_1978} study
the mental operations underlying the comprehension and summarization
of text. They introduce \textit{propositions} as the basic unit of
text representation, and a model of how incoming text is processed
given memory limitations; texts are reduced to important propositions
(to be recalled later) using \textit{macro-operators} (e.g., addition,
deletion). Their model has met with popularity in
cognitive psychology \cite{baddeley07} and has also found application
in summarization \cite{fang-teufel-2016-improving}.

Our work proposes a new encoder for contextualized semantic
parsing. At the heart of our approach is a memory controller which
keeps track of context via writing new information and updating old
information.  Our memory-based approach is inspired by
\citet{kintsch_toward_1978} and is closest to
\citet{pmlr-v48-santoro16}, who use a memory augmented neural network
\citep{DBLP:journals/corr/WestonCB14,NIPS2015_8fb21ee7} for
meta-learning. Specifically, they introduce a method for accessing
external memory which functions as short-term storage for
meta-learning. Although we report experiments solely on semantic
parsing, our encoder is fairly general and could be applied to other
context-dependent tasks such as conversational information seeking
\cite{Dalton:ea:2020} and information retrieval
\cite{SUN2007511,Voorhees:2004}.

%% file: 3_model.tex
\section{Model}
\label{sec:model}
\begin{figure*}[t]
\centering
\includegraphics[width=1\linewidth, keepaspectratio=False]{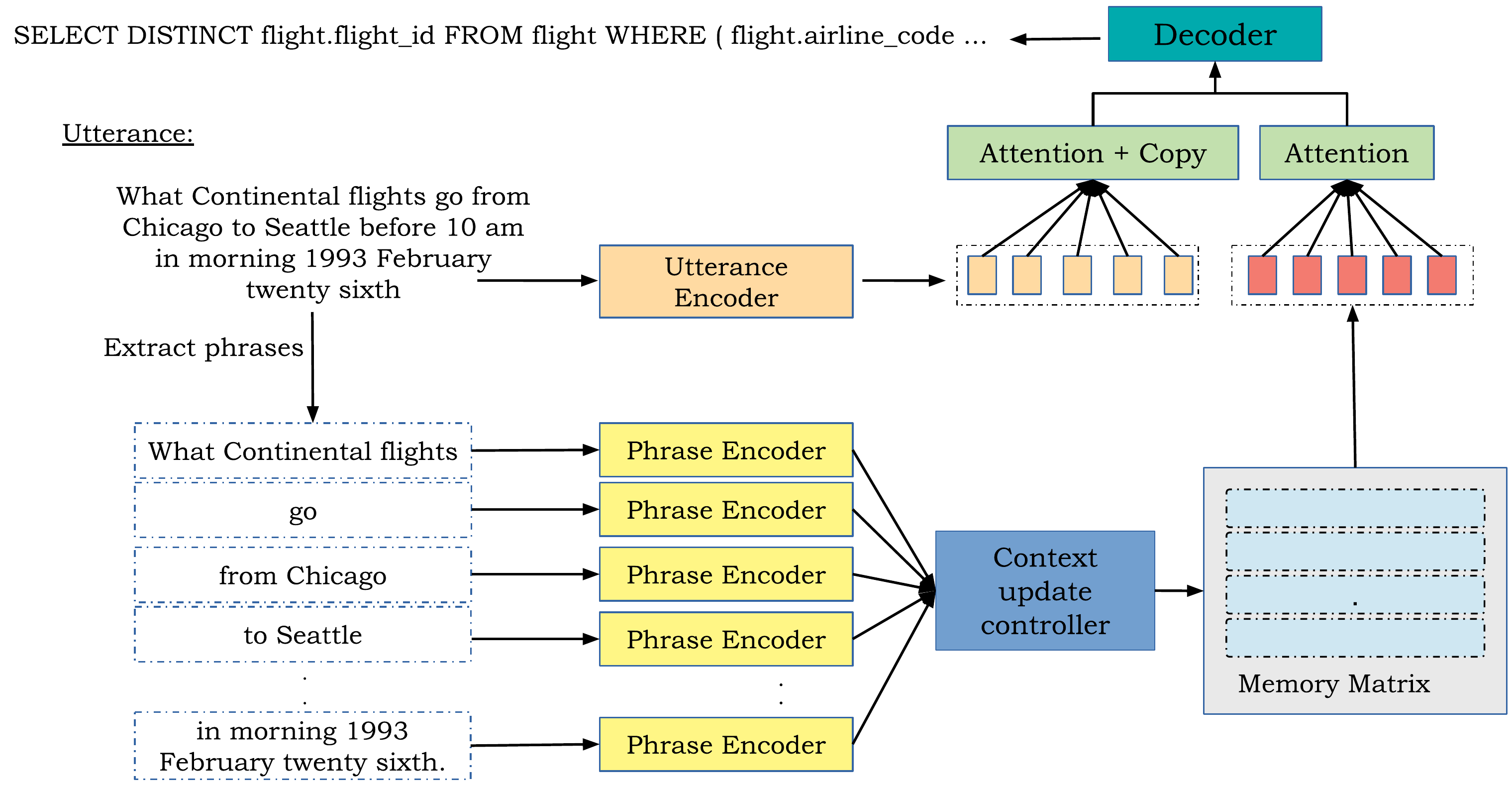}
\caption{Overview of model architecture. Utterances are broken down
  into segments. Each segment is encoded with the same encoder (same
  weights) and is processed independently. The context update
  controller learns to manipulate the memory such that correct
  discourse information is retained.}
\label{fig:system}
\end{figure*}

Our model is based on the encoder-decoder architecture
\citep{Cho_2015} with the addition of a memory component
\citep{NIPS2015_8fb21ee7, pmlr-v48-santoro16} for incorporating
context. Let $I = [X_i, Y_i]_{i=1}^n$ denote an interaction such that
$X_i$ is the input utterance and $Y_i$ is the output SQL at
interaction turn $I[i]$. At each turn~$i$, given~$X_i$ and all
previous turns $I[1 \dots i- 1]$, our task is to predict SQL output~$Y_i$.

As shown in Figure~\ref{fig:system}, our model consists of four
components, (1)~a memory matrix retains discourse information, (2) a
memory controller, learns to access and manipulate the memory such
that correct discourse information is retained, (3) utterance and
phrase encoders, and (4) a decoder which interacts with the memory and
utterance encoder using an attention mechanism to generate SQL output.
\subsection{Input Encoder}
Each input utterance $X_i = (x_{i,1} \dots x_{i,|X_i|})$ is encoded using a bi-directional LSTM~\citep{10.1162/neco.1997.9.8.1735},
\begin{equation}
    h_{i,j}^U = \operatorname{biLSTM}^{U}(e_{i, j}; h^U_{i, j - 1})
    \label{eq:utt_enc}
\end{equation}
where, $e_{i,j} = \phi(x_{i,j})$ is a learned embedding corresponding
to input token $x_{i, j}$ and $h_{i,j}^U$ is the concatenation of the
forward and backward LSTM hidden representations at step~$j$.  As
mentioned earlier, $X_i$~is also segmented into a sequence of phrases
denoted as $X_i = (p_{i}^{1} \dots p_{i}^{K})$, where $K$ is the
number of phrases for utterance $X_i$. We provide details on how
utterances are segmented into phrases in
Section~\ref{sec:experimental-setup}. For now, suffice it to say that
they are obtained from the output of a chunker with some minimal
postprocessing (e.g.,~to merge postmodifiers with NPs or VPs).  Each
phrase consists of tokens $p_{i}^{k} = (x_{i, [s_k: s_k +
  |p_{i}^{k}|]})$, such that $k \in [1,K]$ and $s_k = \sum_{z = 1}^{k
  - 1} |p_{i}^{z}| $. Each phrase $p_{i}^{k}$ is separately
encoded using a bi-directional LSTM,
\begin{equation}
    h_{i, k, j}^P = \operatorname{biLSTM}^{P}(e_{i, j}; h^P_{i, k, j - 1})
    \label{eq:ph_enc}
\end{equation}
such that $j \in [s_k: s_k + |p_{i}^{k}|]$. As shown in
Figure~\ref{fig:system}, every phrase $p_{i}^{k}$ in utterance $i$ is
separately encoded using $\operatorname{biLSTM}^{P}$ to obtain a
phrase representation $h_{i, k}^P$ by concatenating the final forward
and backward hidden representations.


\subsection{Context Memory}
\label{sec:memcontroller}
Our context memory is a matrix $M_i \in \mathbb{R}^{L \times d}$
with~$L$ memory slots, each of dimension~$d$, where~$i$ is the state
of the memory matrix at the $i^{th}$ interaction turn.  The goal of
context memory is to maintain relevant information required to parse
the input utterance at each turn. As shown in Figure~\ref{fig:system},
this is achieved by learning a \emph{context update controller} which
is responsible for updating the memory at each turn.

For each phrase $p_{i}^{k}$ belonging to a sequence of phrases within
utterance $X_i$, the controller decides whether it contains old
information which conflicts with information present in the memory or
new information which has to be added to the current context. When
novel information is introduced, the controller should add it to an
empty or least-used memory slot, otherwise the conflicting memory slot
should be updated with the latest information. Let~$t$ denote the
memory update time step such that $t \in [1, n]$, where $n$ is the
total number of phrases in interaction $I$. We simplify notation,
using~$h_t^P$ instead of~$h_{i, k}^P$, to represent the hidden
representation of a phrase at time~$t$.

\paragraph{Detecting Conflicts} Given phrase
representation~$h_t^P$ (see Equation~\eqref{eq:ph_enc}), we use a
similarity module to detect conflicts between $h_t^P$ and every memory
slot in $M_i(m)$ where $m \in [1, L]$; $M_i(m)$ is the $m^{th}$ row
representing a memory slot in the memory matrix. Intuitively, low
similarity represents new information. Our similarity module is based
on a Siamese network architecture~\citep{NIPS1993_288cc0ff} that takes
phrase hidden representation~$h_t^P$ and memory slot~$M_i(m)$ and
computes a low-dimensional representation using the same neural
network weights. The resulting low-dimensional representations are
then compared using the cosine distance metric:
\begin{equation} 
\label{cos:eq}
\begin{split}
\hspace*{-.7cm}  \hat{w}_c^{t, m}& \hspace*{-1ex}=\hspace*{-.5ex} \frac{\operatorname{sia}(h_t^P) \cdot
    \operatorname{sia}(M_i(m))}{ \max(\parallel
    \operatorname{sia}(h_t^P) \parallel_2 \cdot \parallel
    \operatorname{sia}(M_i(m)) \parallel_2, \epsilon)} \hspace*{-2.5ex}
\end{split}
\end{equation}
where $\epsilon$ is a small value for numerical stability and
$\operatorname{sia}$ is a multi-layer feed-forward network with a
$\tanh$ activation function. 
For hidden representation~$h$, $\operatorname{sia}$ is computed as:
\begin{equation}
\hat{h} = W(\tanh(W^l h + b^l) + b)
\end{equation}
where $l$~represents the layer number and $W^l, b^l, W$, and $b$ are
learnable parameters. We use $\hat{w}_c^{t, m}$ to obtain a similarity
distribution $w_s^{t}$ for updating step~$t$ over memory
slots. $w_s^{t}$ represents the probability of dissimilarity (or
conflict) which is calculated by computing $\operatorname{softmax}$
over cosine similarities with every memory slot~$m \in [1 .. L]$:
\begin{equation}
    w_s^{t} = \operatorname{softmax}([\hat{w}_c^{t, 1};\hat{w}_c^{t, 2} \dots ;\hat{w}_c^{t, L}])
\end{equation}
We compute $\operatorname{softmax}$ over cosine values so that the
linear combination of $w_s^{t}$ with least used weights $w_{lu}^t$ (described below in the memory update paragraph)
still represents the probability of update across each memory slot.

\paragraph{Adding New Information} To add new information to the memory,
i.e.,~when there is no conflict with any locations, we need to
ascertain which memory locations are either empty or rarely used. When
the memory is full, i.e., all memory slots are used during previous
updates, we update the slot which was least used. This is accomplished
by maintaining memory usage weights~$w_u^t \in \mathbb{R}^{L}$ at each
update~$t$; $w_u^t$ is initialized with zeros at $t=0$ and is updated
by combining previous memory usage weights $w_u^{t - 1}$ with current
write weight~$w_w^t$ using a decay parameter~$\lambda$:
\begin{equation}
     w_u^t =   w_w^t + \lambda w_u^{t-1}
\end{equation}
where write weights $w_w^t$ are used to compute the write location and
are described in the memory update paragraph below.
The least used weight vector~$w_{lu}^t$, at update step~$t$ is then
calculated as:
\begin{equation}
    w_{lu}^t  = \operatorname{softmin}(w_{u}^{t - 1})
\end{equation}
where for vector~$x$ we calculate $\operatorname{softmin}(x) =
\exp (-x)/ \sum_{j} \exp (-x_j)$.  Hard
updates, i.e., using $\operatorname{smallest}$ instead of
$\operatorname{softmin}$ are also possible. However, we found
$\operatorname{softmin}$ to be more stable during learning.


\paragraph{Memory Update} \label{sec:memupdate} We wish to compute
write location~$w_w^t$ given least used weight vector~$w_{lu}^t$ and
conflict probability distribution~$w_s^{t}$. Notice that~$w_s^{t}$
and~$w_{lu}^t$ are essentially two probability distributions each
representing a candidate write location in memory. We learn a convex
combination parameter~$\mu$ which depends on~$w_s^{t}$,
\begin{gather}
\label{eq:update}
    \mu = \sigma(W_{\sigma}w_s^{t} + b_{\sigma}) \\
    w_w^{t} = \operatorname{softmax}((\mu w_s^{t} + (1 - \mu)w_{lu}^t)/\tau)
    \label{eq:mem_controller}
\end{gather}
where temperature hyperparameter~$\tau$ is used to peak the write
location. Finally, the memory is updated with current phrase
representation~$h_{t}^P$ as,
\begin{equation}
\hspace*{-.38cm}M_i^t(m)\hspace*{-.4ex}=\hspace*{-.4ex}M_i^{t - 1}(m) +w_w^{t}(m) h_t^P, \forall m \hspace*{-.4ex}\in\hspace*{-.4ex} [1, L]
\label{eq:mem_update}
\end{equation}

\subsection{Decoder}
\label{sec:decoder}
The output query is generated with an LSTM decoder. As shown
in Figure~\ref{fig:system}, the decoder depends on the memory and
utterance representations computed using Equations~(\ref{eq:mem_update})
and~(\ref{eq:utt_enc}), respectively. The decoder state at time step
$s$ is computed as:
\begin{equation}
\hspace*{-.25cm}h_s^D = \operatorname{LSTM}([\phi^o(y_{i,s - 1}); c_{s - 1}^M; c_{s - 1}^U]; h_{s-1}^D)
\end{equation}
where $\phi^o$ is a learned embedding function for output tokens,
$c_s^U$ is an utterance context vector, $c_{s - 1}^M$~is a  memory
context vector, and $h_{s-1}^D$ is the previous decoder hidden
state. $c_s^U$ is calculated as the weighted sum of all hidden states,
where $\alpha_s^U$ is the utterance state attention score:
%
\begin{gather}
    v_s(j) = h_{i,j}^U W{^A} h_s^D \\
    \alpha_s^U = \operatorname{softmax}(v_s) \\
    c_{s}^U = \sum_{j} h_{i,j}^U \alpha_s^U(j)
  \end{gather} 
  {Memory} state attention score $\alpha_s^M$ and {memory} context
  vector~$c_{s}^M$ are computed in a similar manner using memory slots
  as hidden states\footnote{In experiments we found that using the
  (raw) memory directly is empirically better to encoding it with an
  LSTM.}. The probability of output query tokens is computed as:
\begin{gather*}
    P(\hat{w}_{i,s} | X_{i}, Y_{i}, I[:i-1]) \propto \\ \exp(\tanh{([h_s^D;c_{s}^U; c_{s}^M]W^{\hat{o}})}W^o + b^o)
\end{gather*}


We further modify the decoder in order to deal with the large number
of database values (e.g.,~city names) common in text-to-SQL semantic
parsing tasks.  As described in \citet{suhr-etal-2018-learning}, we add
anonymized token attention scores in the output vocabulary
distribution which enables copying anonymized tokens mentioned in
input utterances. The final probability distribution over output
vocabulary tokens and anonymized tokens is:
\begin{gather}
    P(w_{i,s})  = \operatorname{softmax}(P(\hat{w}_{i,s}) \oplus P(\hat{a}_{i,s}))
  \end{gather}
  where $\oplus$ represents concatenation and $P(\hat{a}_{i,s})$ are
  anonymized token attention scores in the attention distribution
  $\alpha_s^U$.

\subsection{Training}
Our model is trained in an end-to-end fashion using a cross-entropy
loss. Given a training set of $N$~interactions $\{I^{(l)}\}_{l=1}^N$,
such that each interaction $I^{(l)}$ consists of utterances $X_i^{(l)}
= (x_{i,1}^{(l)} \dots x_{i,|X_i|^{(l)}})$ paired with output queries
$Y_i^{(l)} = (y_{i,1}^{(l)} \dots y_{i,|Y_i|}^{(l)})$, we minimize
token cross-entropy loss as:
\begin{equation}
    \mathcal{L}(\hat{y}_{i,k}^{(l)}) = -log P(\hat{y}_{i,k}^{(l)} | x_{i}^{(l)}, y_{i,k}^{(l)}, I[:i-1])
\end{equation}
where, $\hat{y}_{i,k}^{(l)}$ denotes the predicted output token
and~$k$ is the gold output token index. The total loss is the
average of the utterance level losses used for back-propagation.




\section{Experimental Setup}
\label{sec:experimental-setup}
We evaluated MemCE, our memory-based context model, on various
settings by integrating it with multiple open-source models.  We
achieve this by replacing the discourse component of related models
with MemCE subject to minor or no additional changes. All base models
in our experiments use a turn-level hierarchical encoder to capture
previous language context.  For primary evaluation, we use the
ATIS~\cite{hemphill-etal-1990-atis,dahl-etal-1994-expanding} dataset
but also present results on SParC~\citep{yu-etal-2019-sparc} and
CoSQL~\citep{yu-etal-2019-cosql}.

 \paragraph{Utterance Segmentation}
\label{sec:chunk}

\begin{figure}
  \begin{center}
\begin{tikzpicture}[
bigsquarednode/.style={rectangle, very thick, minimum size=5mm},
]
\node[bigsquarednode][text width=1\linewidth,align=center]      (stage3)  at (1,1) {[What Continental flights], [go], [from Chicago], [to Seattle], [before 10 am],  [in morning 1993 February twenty sixth.]};

\node[bigsquarednode][text width=1\linewidth,align=center]      (stage2) at (1,3.5) {[What], [Continental flights], [go], [from], [Chicago], [to], [Seattle], [before], [10 am], [in],[morning 1993 February twenty sixth.]
};

\node[bigsquarednode][text width=1\linewidth,align=center]      (stage1) at (1,6)  {What Continental flights go from Chicago to Seattle before 10 am in morning 1993 February twenty sixth.};

\draw[-{Stealth[length=3mm, width=2mm]}] (stage1.south) -- (stage2.north) node [text width=2.5cm,right,midway]{Chunking};
\draw[-{Stealth[length=3mm, width=2mm]}] (stage2.south) -- (stage3.north) node [text width=2.5cm,right,midway]{Merge};

\end{tikzpicture}
\caption{Example of sentence segmentation using chunking and rule-based merging.}
\label{fig:chunk}
\end{center}
\end{figure}
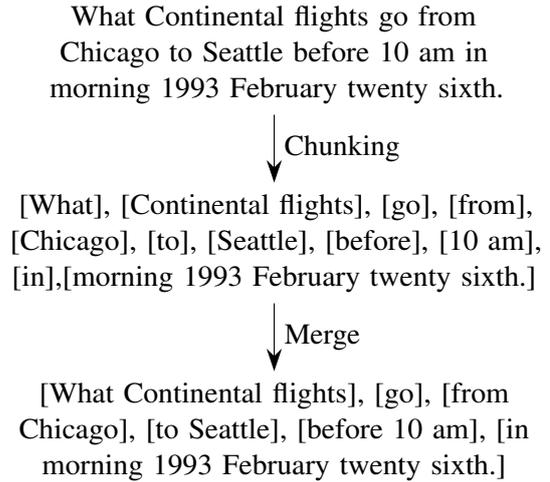

We segment each input utterance into a sequence of phrases with a
pretrained chunker and then apply a simple rule-based merging
procedure to create bigger chunks as an approximation to propositions
\cite{kintsch_toward_1978}. Figure~\ref{fig:chunk} illustrates the
process.  We used the Flair chunker~\citep{akbik-etal-2018-contextual}
trained on
Conll-2000~\citep{tjong-kim-sang-buchholz-2000-introduction} to
identify NP and VP phrases without postmodifiers. Small chunks
(e.g.,~\textsl{from, before} in the figure) were subsequently merged
into segments using the following rules and
NLTK's~\cite{bird2009natural} tag-based regex merge:

\begin{fleqn}
\begin{equation*}
\begin{alignedat}{2}
\text{R1: }&left = \langle VP.*\rangle, right =\langle VP.*\rangle\\
\text{R2: }&left =\langle PP.*\rangle|\langle NP.*\rangle, right =\langle NP\rangle+\\
\text{R3: }&left= \langle NP.*\rangle, right =\langle VB.*\rangle\\
\text{R4: }&left= \langle AD.*\rangle, right= \langle NP.*\rangle
\end{alignedat}
\end{equation*}
\end{fleqn}

The rules above are applied in order. For each rule we find any chunk
whose end matches the left pattern followed by a chunk whose beginning
matches the right pattern. Chunks that satisfy this criterion are
merged.

We segment utterances and anonymize entities independently and then
match entities within segments deterministically. This step is
necessary to robustly perform anonymization as in some rare cases, the
chunking process will separate entities in two different phrases
(e.g., \textsl{in Long Beach California that} is chunked as \textsl{in
  Long Beach} and \textsl{California that}). This is easily handled by
a simple token number matching procedure between the anonymized
utterance and corresponding phrases.

\paragraph{Model Configuration}
Our model is implemented in PyTorch~\citep{NEURIPS2019_bdbca288}. For
all experiments, we used the ADAM optimizer~\citep{kingma2019method}
to minimize the loss function and the initial learning rate was set
to~0.001. During training, we used the
$\operatorname{ReduceLROnPlateau}$ learning rate scheduling strategy
on the validation loss, with a decay rate of~0.8. We also applied
dropout with 0.5~probability. Dimensions for the word embeddings were
set to~300. Following previous work \citep{zhang-etal-2019-editing} we
use pretrained GloVe~\citep{pennington-etal-2014-glove} embeddings for
our main experiments on the SparC and CoSQL datasets. For ATIS, word
embeddings were not pretrained
\citep{suhr-etal-2018-learning,zhang-etal-2019-editing}. Memory length
was chosen as a hyperparameter from the range~\mbox{[15, 25]} and the
temperature parameter was chosen from \{0.01, 0.1\}. Best memory
length values for ATIS, SparC, and CoSQL were 25, 16, and 20,
respectively. The RNN decoder is a two-layer LSTM and the encoder is a
single layer LSTM. The Siamese network in the module which detects
conflicting slots uses two hidden layers.

%% file: 4_results.tex
\begin{table*}[ht]
\centering
\resizebox{0.9\textwidth}{!}{
\begin{tabular}{@{\extracolsep{8pt}}lccccccc@{}}
\Xhline{4\arrayrulewidth}
      \multirow{3}{*}{Model} & \multirow{3}{*}{Enc-Dec} & \multicolumn{3}{c}{Dev Set}   &    \multicolumn{3}{c}{Test Set} \\ \cline{3-5} \cline{6-8} 
      & & & \multicolumn{2}{c}{Denotation}   &    &    \multicolumn{2}{c}{Denotation} \\ \cline{4-5} \cline{7-8} 
 & &Query & Relaxed & Strict & Query & Relaxed & Strict \\ \hline
Seq2Seq & LSTM-LSTM & 28.7 & 48.8 & 43.2 & 35.7 & 56.4 & 53.8\\
Seq2Seq+Concat & LSTM-LSTM& 35.1 & 59.4 & 56.7 & 42.2 & 66.6 & 65.8 \\
\citet{suhr-etal-2018-learning} & HE-LSTM   & 36.0 & 59.5 & 58.3   & --- & --- & --- \\
\citet{suhr-etal-2018-learning} & HE-SnipCopy   & 37.5 & 63.0 & 62.5  & 43.6 & 69.3 & 69.2 \\
\citet{zhang-etal-2019-editing}& HE-EditBased  & 36.2 & 60.5 & 60.0 & 43.9 & 68.5 & 68.1 \\ 
\citet{Lin2019GrammarbasedNT}& LSTM-Grammar  & 39.1 & --- & 65.8 & 44.1 & --- & 73.7 \\ \hdashline
MemCE & Mem-LSTM& 40.2 & 63.6 & 61.2 & 47.0 & 70.1 & 68.9 \\
MemCE & Mem-SnipCopy & 39.1 & 65.5 & 65.2 & 45.3 & 70.2 & 69.8 \\
\Xhline{4\arrayrulewidth}
\end{tabular}
}
\caption{Model accuracy on the ATIS dataset. HE is
  a hierarchical interaction encoder,  while Mem is the proposed
  memory-based encoder. LSTM are vanilla encoder/decoder models, while
  SnipCopy copies SQL segments from the previous query and EditBased
  adopts a query editing mechanism.}
\label{tab:atis_result}
\end{table*}




\begin{table*}[ht]
\centering
\resizebox{1\textwidth}{!}{
\begin{tabular}{@{\extracolsep{0pt}}lccccccccc|cc@{}}
\Xhline{4\arrayrulewidth}
     \multirow{2}{*}{Model} & \multirow{2}{*}{Enc-Dec} & \multicolumn{2}{c}{CoSQL(D)} & \multicolumn{2}{c}{CoSQL(T)}   &  \multicolumn{2}{c}{SparC(D)} & \multicolumn{2}{c}{SparC(T)} & \multicolumn{2}{c}{SparC-DI(T)} \\ \cline{3-4} \cline{5-6} \cline{7-8} \cline{9-10} \cline{11-12}
 &  & Q & I & Q & I & Q & I & Q & I & Q & I \\ \hline
 CDS2S & HE-LSTM  & 13.8 & 2.1 & 13.9 & 2.6 & 21.9  & 8.1 & 23.2 & 7.5 & 39.5  & 20.1\\
 CDS2S & HE-SnipCopy & 12.3 & 2.1 & --- & ---   & 21.7  & 9.5 & 20.3 & 8.1 & 38.7  & 24 \\ 
 \citet{ijcai2020-495} & HE-Grammar & 33.5 & 9.6 & --- & ---  & 41.8 & 20.6 & --- & --- & 57.1 & 35.3\\ \hdashline
 MemCE+CDS2S & Mem-LSTM & 13.4 & 3.4 & --- & ---  & 21.2  & 8.8  & --- & --- & 41.3  & 22.9 \\
 MemCE+CDS2S & Mem-SnipCopy & 13.1 & 2.7 & --- & ---   & 21.4  & 10.9 & --- & --- & 41.5  & 26.7\\
 MemCE+\citet{ijcai2020-495} & Mem-Grammar & 32.8 & \hspace*{-.2cm}10.6 & 28.4  & 6.2 & 42.4 & 21.1 & 40.3 & 16.7 & 55.7 & 36.3 \\ 

\Xhline{4\arrayrulewidth}
\end{tabular}
} 
\caption{Query (Q) and Interaction (I) accuracy for
  SParC and CoSQL. We report results on the development (D) and test (T)
  sets. Sparc-DI is our domain-independent split of
  SparC. HE is a hierarchical encoder and Mem is the
  proposed memory-based context encoder. LSTM is a vanilla decoder,
  SnipCopy copies SQL segments from the previous query, and Grammar
  refers to a decoder which outputs a sequence of grammar rules rather
  than tokens. Table cells are filled with --- whenever results are not
  available.}  
\label{tab:sparc_result}
\end{table*}

\section{Results}
In this section, we assess the effectiveness of the MemCE encoder at
handling contextual information. We present our results, evaluation
methodology, and comparisons against the state of the art.
\subsection{Evaluation on ATIS}
\label{sec:eval-atis}

We primarily focus on ATIS because it contains relatively long
interactions (average length is~7) compared to other datasets
(e.g,~the average length in SParC is~3). Longer interactions present
multiple challenges that require non-trivial processing of context,
some of which are discussed in Section~\ref{sec:analysis}. We use the
ATIS dataset split created by~\citet{suhr-etal-2018-learning}. It
contains 27 tables and 162K entries with 1,148/380/130 train/dev/test
interactions. The semantic representations are in SQL.


Following \citet{suhr-etal-2018-learning}, we measure \emph{query
  accuracy}, \emph{strict denotation accuracy}, and \emph{relaxed
  denotation accuracy}. Query accuracy is the percentage of predicted
queries that match the reference query. Strict denotation accuracy is
the percentage of predicted queries that when executed produce the
same results as the reference query. Relaxed accuracy also gives
credit to a prediction query that fails to execute if the reference
table is empty. In cases where the utterance is ambiguous and there
are multiple gold queries, the query or table is considered correct if
they match any of the gold labels. We evaluate on both development and
test set, and select the best model during training via a separate
validation set consisting of $5\%$ of the training data.


Table~\ref{tab:atis_result} presents a summary of our results. We
compare our approach against a simple Seq2Seq model which is a
baseline encoder-decoder without any access to contextual
information. Seq2Seq+Concat is a strong baseline which consists of an
encoder-decoder model with attention on the current and the
\emph{previous three concatenated} utterances.  We also compare
against the models of~\citet{suhr-etal-2018-learning}
and~\citet{zhang-etal-2019-editing}. The former employs a turn-level
encoder on top of an utterance-level encoder in a \emph{hierarchical}
fashion together with a decoder which learns to copy complete SQL
segments from the previous query (SQL segments between consecutive
queries are aligned during training using a rule-based procedure).
The latter enhances the turn-level encoder by employing an attention
mechanism across different turns and additionally introduces a
\emph{query editing} mechanism which decides at each decoding step
whether to copy from the previous query or insert a new token.  Column
Enc-Dec in Table~\ref{tab:atis_result} describes the various models in
terms of the type of encoder/decoder used. LSTM is a vanilla encoder
or decoder, HE is a turn-level hierarchical encoder, and Mem is the
proposed memory-based encoder.  SnipCopy and EditBased respectively
refer to Suhr et al.'s \shortcite{suhr-etal-2018-learning} and Zhang
et al.'s \shortcite{zhang-etal-2019-editing} decoders.  We present two
instantiations of our MemCE model with a simple LSTM decoder
(Mem-LSTM) and SnipCopy (Mem-SnipCopy). For the sake of completeness,
Table~\ref{tab:atis_result} also reports the results from
\citet{Lin2019GrammarbasedNT} who apply a grammar-based decoder to
this task; they also incorporate the interaction history by
concatenating the current utterance with the previous three utterances
which are encoded with a bi-directional LSTM. All models in
Table~\ref{tab:atis_result} use entity anonymization,
\citet{Lin2019GrammarbasedNT} additionally use identifier linking,
i.e.,~string matching heuristic rules to link words or phrases in the
input utterance to identifiers in the database
(e.g.,~\texttt{city\_name\_string -> ``BOSTON''}).

As shown in Table~\ref{tab:atis_result},  MemCE is able to outperform
comparison systems. We observe a boost in denotation accuracy
when using the SnipCopy decoder instead of an LSTM-based one, however,
exact match does not improve.  This is possibly because SnipCopy makes
it easier to generate long SQL queries by copying segments, but at the
same time it suffers from spurious generation and error propagation.

Table~\ref{tab:atis-ablation} presents various
  ablation studies which evaluate the contribution of individual model
  components. We use Mem-SnipCopy as our base model and report
  performance on the ATIS development set following the configuration
  described in Section~\ref{sec:experimental-setup}. We first remove
  the proposed memory controller described in
  Section~\ref{sec:memcontroller} and simplify
  Equation~\eqref{eq:mem_controller} using key-value based attention
  to  calculate $w_w^t$ as,
\begin{gather}
\label{eq:updateb}
    \alpha_{j} = M_{i}^{t-1}(j) W{^P} h_t^P \\
    w_{w}^{t} = \operatorname{softmax}(\alpha)
\end{gather}
We observe a decrease in performance (see second row in
Table~\ref{tab:atis-ablation}) indicating that the proposed memory
controller is helpful in maintaining interaction context.

We performed two ablation experiments to evaluate the usefulness of
utterance segmentation. Firstly, instead of the phrases extracted from
our segmentation procedure, we employ a variant of our model which
operates over individual tokens (see row ``phrases are utterance
tokens'' in Table~\ref{tab:atis-ablation}). As can be seen, this
strategy is not optimal as results decrease across metrics. We believe
operating directly on tokens can lead to ambiguity during update. For
example, when processing current phrase \textsl{to Boston} given
previous utterance \textsl{What Continental flights go from Chicago to
  Seattle}, it is not obvious whether \textsl{Boston} should update
\textsl{Chicago} or \textsl{Seattle}.  Secondly, we do not use any
segmentation at all, not even at the token level. Instead, we treat
the entire utterance as a single phrase (see row ``phrases are full
utterances'' in Table~\ref{tab:atis-ablation}). If memory's only
function is to simply store utterance encodings, then this model
becomes comparable to a hierarchical encoder with attention. Again, we
observe that performance decreases which indicates that our system
benefits from utterance segmentation. Overall, the ablation studies in
Table~\ref{tab:atis-ablation} show that segmentation and its
granularity matters. Our heuristic procedure works well for the task
at hand, although a learning-based method would be more flexible and
potentially lead to further improvements. However, we leave this to
future work.


\begin{table}[t]
\centering
\resizebox{1\linewidth}{!}{
\begin{tabular}{@{}lccc@{}}
  \toprule
  & & \multicolumn{2}{c}{Denotation} \\
  \cline{3-4}
  & Query & Relaxed & Strict \\ \hline
  MemCE+SnipCopy & 39.1 & 65.5 & 65.2  \\
\hspace{0.2cm} Without memory controller  & 34.3 & 58.7 & 58.1  \\
\hspace{0.2cm} Phrases are utterance tokens  & 37.2 & 61.9 & 61.7\\
\hspace{0.2cm} Phrases are full utterances  & 36.8 & 64.2 & 63.9  \\
  \bottomrule
\end{tabular}
}
\caption{Ablation results with SnipCopy decoder on the ATIS
  development set.}
\label{tab:atis-ablation}
\end{table}

\begin{table}[t]
\centering
\small
\begin{tabular}{lccc}
\toprule
               & Train & Dev & Test \\ \hline
\#Interactions & 2869  & 290 & 290  \\
\#Utterances   & 8535  & 851 & 821 \\
\bottomrule
\end{tabular}
\caption{Statistics for SParC-DI domain-independent split 
  which has 157~domains in total.} 
\label{tab:sparc-resplit}
\end{table}

%
%
%
%
%


\subsection{Evaluation on SParC and CoSQL}
\label{sec:eval-sparc}

In this section we describe our results on SParC and CoSQL. Both
datasets assume a cross-domain semantic parsing task in context with
SQL as the meaning representation. In addition, for ambiguous
utterances, (which cannot be uniquely mapped to SQL given past
context) CoSQL also includes clarification questions (and answers). We
do not tackle these explicitly but consider them part of the utterance
preceding them (e.g., \textsl{please list the singers} | \textsl{did
  you mean list their names?} | \textsl{yes}). Since our primary
objective is to study and measure context-dependent language
understanding, we created a split of SParC which is denoted as
SParC-DI\footnote{We only considered training and development
instances as the test set is not publicly available.}  where domains
are all seen in training, development, \emph{and} test set. In this
way we ensure that no model has the added advantage of being able to
handle cross-domain instances while lacking context-dependent language
understanding. Table~\ref{tab:sparc-resplit} shows the statistics of
our SParC-DI split, following a ratio of 80/10/10 percent for the
training/development/test set.


We evaluate model output using exact set match
accuracy~\citep{yu-etal-2019-sparc}.\footnote{Predicted queries are
  decomposed into different SQL clauses and scores are computed for
  each clause separately.} We report two metrics: \textit{question
  accuracy} which is the accuracy considering all utterances
independently, and \textit{interaction accuracy} which is the correct
interaction accuracy averaged across interactions. An interaction is
marked as correct if all utterances in that interaction are
correct. Since utterances in an interaction can be semantically
complete (i.e., independent of context), we prefer interaction
accuracy.


Table~\ref{tab:sparc_result} summarizes our results.  CDS2S is the
context-dependent cross-domain parsing model of
\citet{zhang-etal-2019-editing}. It is is adapted
from~\citet{suhr-etal-2018-learning} to include a schema encoder which
is necessary for SparC and CoSQL. It also uses a turn-level
hierarchical encoder to represent the interaction history. We also
report model variants where the CDS2S encoder is combined with an
LSTM-based encoder, SnipCopy \cite{suhr-etal-2018-learning} and a
grammar-based decoder \citet{ijcai2020-495}. The latter  decodes SQL
queries as a sequence of grammar rules, rather than tokens. We compare
the above systems with three variants of our MemCE model which differ
in their use of an LSTM decoder, SnipCopy, and the Grammar-based
decoder of \citet{ijcai2020-495}.

Across models and datasets we observe that MemCE improves performance
which suggests that it better captures contextual information as an
independent language modeling component. We observe that benefits from
our memory-based encoder persist across domains and data splits even
when sophisticated strategies like grammar-based decoding are adopted.


%% file: 5_analysis.tex
\definecolor{h1}{HTML}{d8b365}
\definecolor{h2}{HTML}{5ab4ac}

\begin{figure*}[t]
\centering
\includegraphics[width=0.95\linewidth, keepaspectratio=true]{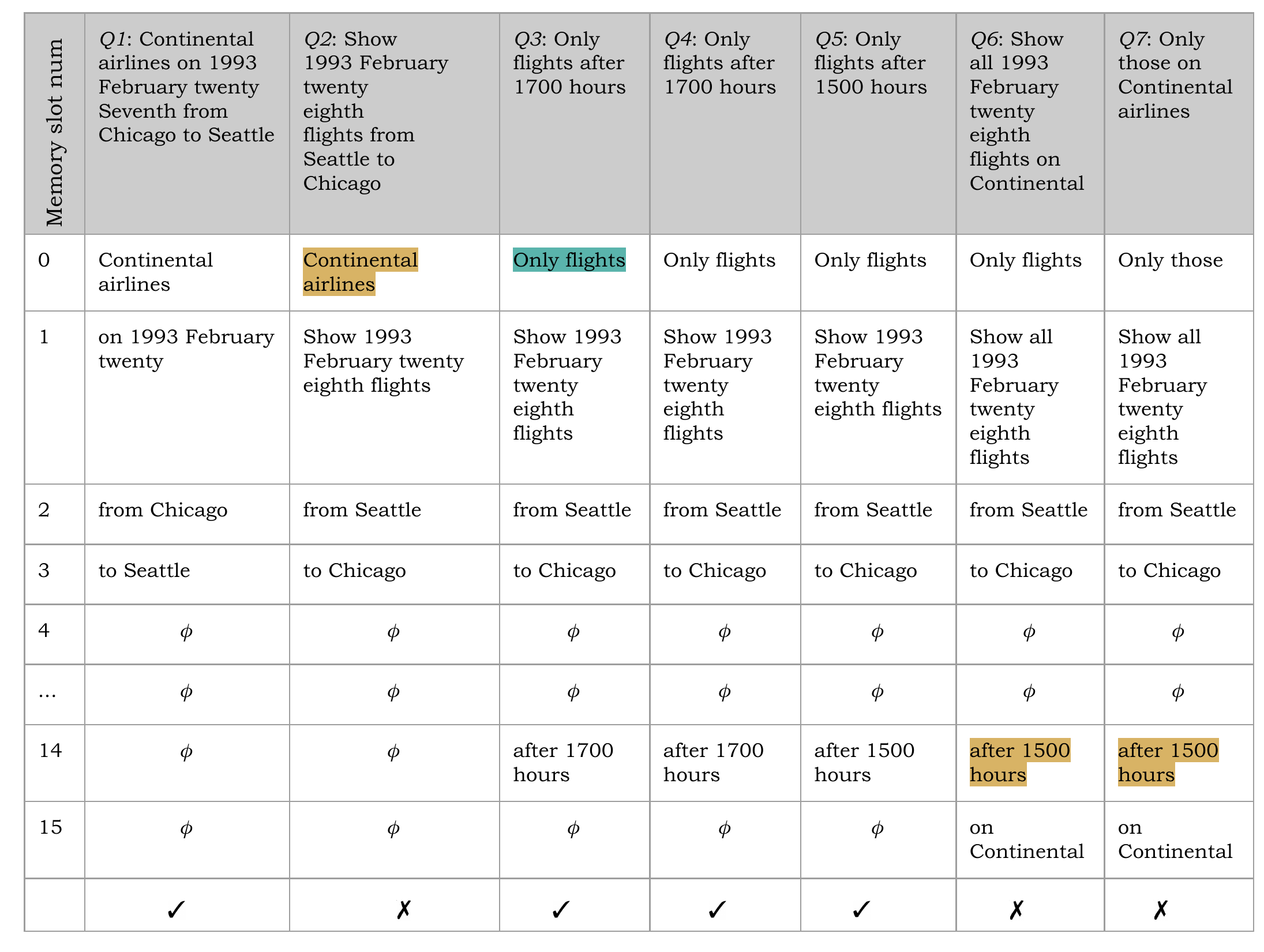}
\caption{Visualization of memory matrix. Rows represent memory content
  and columns represents the utterance time step. The top row shows
  the utterances being processed. Each row is marked with a memory
  slot number which represents the content of memory in that
  slot. Empty slots are marked with $\phi$. The bottom row shows
  whether the utterance was parsed correctly(\ding{51}) or
  not(\ding{55}). \fcolorbox{black}{h1}{\rule{0pt}{5pt}\rule{5pt}{0pt}}:
  Stale content in memory w.r.t the current
  utterance. \fcolorbox{black}{h2}{\rule{0pt}{5pt}\rule{5pt}{0pt}}:
  Incorrect substitution.}
\label{fig:mem}
\end{figure*}

\begin{table}[t]
\centering
\small
\resizebox{1\linewidth}{!}{
\begin{tabular}{lcccc}
  \toprule
  & \multicolumn{2}{c}{MemCE} & \multicolumn{2}{c}{Suhr et al. (2018)} \\ 
  \cline{2-3} \cline{4-5}
  & Denotation & Query & Denotation & Query \\ \hline
  Focus Shift & 80.4 & 50.0 & 76.7 & 44.6  \\
  Referring Exp & 80.0 & 40.0 & 70.0 & 20.0  \\
  Ellipsis & 69.4 & 33.3 &  66.6 & 25.0  \\
  Independent & 81.4 & 61.1 & 81.3 & 62.7  \\
  \bottomrule
\end{tabular}
}
\caption{Model accuracy on specific phenomena  (20 interactions, ATIS dev set).} 
\label{tab:analysis}
\end{table}

\section{Analysis}
\label{sec:analysis}
In this section, we analyze our model's ability to handle important
discourse phenomena such as focus shift, referring expressions, and
ellipsis.  We also showcase its interpretability by examining the
behavior of the (learned) memory controller.


\subsection{Focus Shift} 
Our linguistic analysis took place on 20
interactions\footnote{Interactions with less than two utterances were
discarded.} randomly sampled from the ATIS development set
(134~utterances in total). Table~\ref{tab:analysis} shows overall
performance statistics for MemCE (Mem-LSTM) and
\citet{suhr-etal-2018-learning} (HE-SnipCopy) on our sample.  We
annotated the focus of attention in each utterance (underlined in the
example below) which we operationalized as the most salient entity
(e.g., city) within the utterance
\cite{grosz-etal-1995-centering}. Focus shift occurs when the
attention transitions from one entity to another. In the interaction
below the focus shifts from \textsl{flights} in Q2 to \textsl{cities}
in Q3. 
\begin{tcolorbox}[colback=white]
\small\vspace*{-.2cm}
\hspace*{-.48cm}Q1: What \uline{flights} are provided by American airlines\hspace*{-1cm} \\
\hspace*{-.48cm}Q2: What \uline{flights} are provided by Delta airlines \\
\hspace*{-.48cm}Q3: Which \uline{cities} are serviced by both American \hspace*{.2cm}and Delta airlines\vspace*{-.2cm}
\end{tcolorbox}

Handling focus shift has been problematic in the context of semantic
parsing \cite{suhr-etal-2018-learning}. In our sample, 41.8\% of
utterances displayed focus shift. Our model was able to correctly
parse all utterances in the interaction above and is more apt at
handling focus shifts compared to related systems
\cite{suhr-etal-2018-learning}.  Table~\ref{tab:analysis} reports
denotation and query accuracy on our analysis sample.


\subsection{Referring Expressions and Ellipsis} 
\label{ling-analysis-coref}

Ellipsis refers to the omission of information from an utterance that
can be recovered from the context. In the interaction below, Q2 and Q3
exemplify nominal ellipsis, the NP \textsl{all flights from Long Beach
  to Memphis} is elided and ideally should be recovered from the
discourse, in order to generate correct SQL queries. Q4 is an example
of coreference, \textsl{they} refers to the answer of Q3. However, it
can also be recovered by considering all previous utterances
(i.e.,~Where do {they} [flights from Long Beach to Memphis; any day]
stop).  Since our model explicitly stores information in context, it
is able to parse utterances like Q2 and Q4 correctly.

\begin{tcolorbox}[colback=white]
\small
\vspace*{-.2cm}
\hspace*{-.48cm}Q1: Please give me \uline{all flights from Long Beach to} \hspace*{.2cm}\uline{Memphis}\\
\hspace*{-.48cm}Q2: What {about} 1993 June thirtieth\\
\hspace*{-.48cm}Q3: How {about} any day \\
\hspace*{-.48cm}Q4: Where do \uline{they} stop \vspace*{-.2cm}
\end{tcolorbox}

In our ATIS sample, 26.8\%~of the utterances exhibited ellipsis and
7.5\%~contained referring expressions. Results in
Table~\ref{tab:analysis} show that MemCE is able to better handle both
such cases. 

\subsection{Memory Interpretation}
\label{mem-analysis}
In this section we delve into the memory controller with the aim of
understanding what kind of patterns it learns and where it fails.  In
Figure~\ref{fig:mem}, we visualize the content of memory for an
interaction (top row) from the ATIS development set consisting of
seven utterances.\footnote{Q4 was repeated in the dataset. We do the
  same to maintain consistency and to observe the effect of
  repetition.}  Each column in Figure~\ref{fig:mem} shows the content
of memory after processing the corresponding utterance in the
interaction. The bottom row indicates whether the final output was
correct (\ding{51}) or not (\ding{55}). For the purpose of clear
visualization we took the $\operatorname{max}$ instead of
$\operatorname{softmax}$ in Equation~\eqref{eq:update} to obtain the
memory state at any time step.




Q2 presents an interesting case for our model, it is not obvious
whether \textsl{Continental airlines} from Q1 should be carried
forward while processing Q2. The latter is genuinely ambiguous, it
could be referring to Continental airlines flights or to flights by
any carrier leaving from Seattle to Chicago.  If we assume the second
interpretation, then Q2 is more or less semantically complete and
independent of Q1.  44\% of  utterances in our ATIS sample are
semantically complete. Although we do not explicitly handle such
utterances, our model is able to parse many of them correctly because
they usually repeat the information mentioned in previous discourse as
a single query (see Table~\ref{tab:analysis}).  Q2 also shows that the
memory controller is able to learn the similarity between long
phrases: \textsl{on 1993 February twenty Seventh} $\Leftrightarrow$
\textsl{Show 1993 February twenty eighth flights}. It also demonstrates
a degree of semantic understanding, i.e., it replaces 
\textsl{from Chicago} with \textsl{from Seattle} in order to process
utterance Q2, rather than simply relying on entity matching.


Figure~\ref{fig:mem} further shows the kind of mistakes the controller
makes which are mostly due to stale content in memory. In utterance~Q6
the memory carries over the constraint \textsl{after 1500 hours} from
the previous utterance which is not valid since Q6 explicitly states
\textsl{Show all $\dots $flights on Continental}. At the same time
constraints \textsl{from Seattle} and \textsl{to Chicago} should carry
forward. Knowing which content to keep or discard makes the task
challenging.

Another cause of errors relates to reinstating previously nullified
constraints. In the interaction below, Q3 reinstates \textsl{from
  Seattle to Chicago}, the focus shifts from flights in Q1 to ground
transportation in Q2 and then again to flights in Q3.
\begin{tcolorbox}[colback=white]
\small \vspace*{-.2cm}\hspace*{-.48cm}Q1: Show flights \uline{from Seattle to Chicago} \\
\hspace*{-.48cm}Q2:\hspace*{.1cm}What ground transportation is
available in \hspace*{.2cm}Chicago \\
\hspace*{-.48cm}Q3: Show flights after 1500 hours\vspace*{-.2cm}
\end{tcolorbox}

Handling these issues altogether necessitates a non-trivial way of
managing context. Given that our model is trained in an end-to-end
fashion, it is encouraging to observe a one-to-one correspondence
between memory and the final output which supports our hypothesis that
explicitly modeling language context is helpful.

%% file: 7_conclusion.tex
\section{Conclusions}
\label{sec:conclusions}

In this paper, we presented a memory-based model for context-dependent
semantic parsing and evaluated its performance on a text-to-SQL
task. Analysis of model output revealed that our approach is able to
handle several discourse related phenomena to a large extent. We also
analyzed the behavior of the memory controller and observed that it
correlates with the model's output decisions. Our study indicates that
explicitly modeling context can be helpful for contextual language
processing tasks. Our model manipulates information at the phrase
level which can be too rigid for fine-grained updates. In the future,
we would like to experiment with learning the right level of utterance
segmentation for context modeling as well as learning when to
reinstate a constraint.
\section*{Acknowledgement}
We thank Mike Lewis, Miguel Ballesteros, and our anonymous reviewers for their feedback. We are grateful to Alex Lascarides and Ivan Titov for their comments on the paper. This work was supported in part by Huawei and the UKRI Centre for Doctoral Training in Natural Language Processing (grant EP/S022481/1). Lapata acknowledges the support of the European
Research Council (award number 681760, ``Translating Multiple
Modalities into Text'').

